\documentclass{article}
\usepackage{ijcai24}

\usepackage{times}
\usepackage{soul}
\usepackage{url}
\usepackage[hidelinks]{hyperref}
\usepackage[utf8]{inputenc}
\usepackage[small]{caption}
\usepackage{graphicx}
\usepackage{amsmath}
\usepackage{amsthm}
\usepackage{booktabs}
\usepackage{algorithm}
\usepackage{algorithmic}
\usepackage[switch]{lineno}

\usepackage{amsmath}
\usepackage{makecell}
\usepackage{array}
\usepackage{multirow}

 \usepackage{enumitem}
\usepackage{color,soul}
\usepackage{amsfonts,amssymb}
\usepackage[switch]{lineno}
\usepackage[dvipsnames]{xcolor}

\newcommand{\ie}{\textit{i}.\textit{e}.}
\newcommand{\eg}{\textit{e}.\textit{g}.}

\newcommand{\whzjh}[1]{{\color{black}#1}}

\title{Shap-Mix: Shapley Value Guided Mixing for Long-Tailed \\ Skeleton Based Action Recognition}

\author{
Jiahang Zhang
\and
Lilang Lin\and
Jiaying Liu
\thanks{Corresponding author.}
\affiliations
Wangxuan Institute of Computer Technology, Peking University\\
\emails
\{zjh2020, linlilang, liujiaying\}@pku.edu.cn
}

\begin{document}

\maketitle

\begin{abstract}
In real-world scenarios, human actions often fall into a long-tailed distribution. It makes the existing skeleton-based action recognition works, which are mostly designed based on balanced datasets, suffer from a sharp performance degradation. Recently, many efforts have been made to image/video long-tailed learning. However, directly applying them to skeleton data can be sub-optimal due to the lack of consideration of the crucial spatial-temporal motion patterns, especially for some modality-specific methodologies such as data augmentation. To this end, considering the crucial role of the body parts in the spatially concentrated human actions, we attend to the mixing augmentations and propose a novel method, Shap-Mix, which improves long-tailed learning by mining representative motion patterns for tail categories. Specifically, we first develop an effective spatial-temporal mixing strategy for the skeleton to boost representation quality. Then, the employed saliency guidance method is presented, consisting of the saliency estimation based on Shapley value and a tail-aware mixing policy. It preserves the salient motion parts of minority classes in mixed data, explicitly establishing the relationships between crucial body structure cues and high-level semantics. Extensive experiments on three large-scale skeleton datasets show our remarkable performance improvement under \textit{both long-tailed and balanced} settings. Our project is publicly available at: {https://jhang2020.github.io/Projects/Shap-Mix/Shap-Mix.html}.
\end{abstract}

\section{Introduction}
\label{sec:intro}
Human activity understanding is a crucial problem with wide applications in real life, \eg, healthcare and autonomous driving.
3D skeleton, as a highly efficient representation, describes the human body by 3D coordinates of keypoints. In comparison to other modalities such as RGB videos and depth data, skeletons are lightweight, compact, and relatively privacy-preserving. Owing to these advantages, skeletons have become widely used in human action recognition.

\begin{figure}[t]
    \centering
    \includegraphics[width=0.48\textwidth]{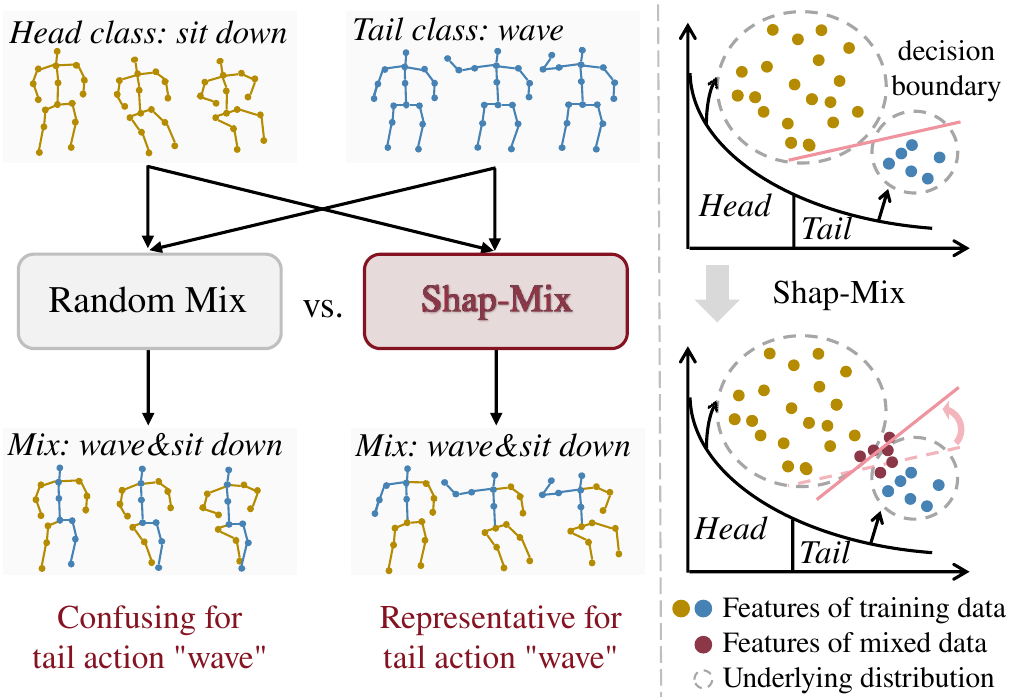}
   \caption{
   Random mix, \eg, Cut-Mix, treats different classes equally and causes the semantic confusion, degrading the long-tailed performance especially for tail categories. In contrast, our Shap-Mix generates representative samples for tail categories to recover the underlying distribution, obtaining a better decision boundary.
   }
   \vspace{-5pt}
  \label{fig:teaser}
\end{figure}

Skeleton-based human action recognition has achieved a remarkable success~\cite{song2018skeleton,chen2021channel,zhang2022hiclr}. However, existing works are mostly targeted at balanced datasets, ignoring the prevalent long-tailed distribution phenomenon in the real world. For example, NTU datasets~\cite{shahroudy2016ntu,liu2019ntu}, currently the most popular benchmarks for skeleton-based action recognition, are constructed with a balanced data distribution obtained by human intervention. In contrast, the human actions in real world are under long-tailed distributions~\cite{zhang2021videolt,Damen2022RESCALING}. For example, there are lots of data on actions of ``walking" or ``standing" while little on some actions such as ``abdominal pain" and ``falling", which can be more important for some real applications, \eg, health monitoring. This gap results in a huge limitation for the deployment of these works in the real world, \ie, a significant performance degradation on tail categories when transferred to the long-tailed settings directly, due to the triggered biased training~\cite{liu2023balanced}.

To deal with this prevalent challenge in real world, long-tailed learning has attracted much attention. Generally, the training data under the long-tailed distribution brings two challenges to model~\cite{chu2020feature}. The first challenge is the \textit{data imbalance}, which can cause representation bias and negative gradient over-suppression problem on the tail classes~\cite{hsieh2021droploss,tan2020equalization}. To this end, many re-balancing methods have been proposed, including the re-sampling~\cite{van2007experimental,han2005borderline,shen2016relay} and re-weighting~\cite{lin2017focal,ren2020balanced,park2021influence}, to help the model learn a balanced representation space. They are built upon the assumption that the optimal decision boundary is still well-defined in the given partial data. However, it is often difficult to obtain the optimal decision boundary due to the scarce tail class data. This introduces the second challenge, \ie, \textit{the difficulty of recovering the underlying distributions} from limited tail class samples. As an effective way to provide new training samples, data augmentation~\cite{chu2020feature,chou2020remix,chen2022imagine} has been widely studied to enhance the distribution learning for the tail categories. However, previous augmentation-based long-tailed methods are designed for images/videos and cannot directly transfer to the skeleton data, which necessitates more careful consideration of its crucial spatial-temporal motion patterns and has been overlooked before. Meanwhile, to get rid of the complicated multiple training stages in previous works~\cite{chu2020feature,chen2022imagine}, we also aim to develop a simpler augmentation-based method with end-to-end training.

In this paper, we focus on the long-tailed augmentation-based method. Considering that human action is often represented as the combination of different motion patterns in skeletons, we attend to the mixing method to integrate different motion parts. Specifically, we propose a novel method, Shap-Mix, as shown in Figure~\ref{fig:teaser}. It can generate representative samples for the tail categories guided by the Shapley value~\cite{shapley1953value}, a important solution concept for allocation problems in cooperative game theory. To begin with, a simple random mixing technique is developed with effective spatial-temporal design, significantly improving the representation quality of the encoder backbone. Based on this, considering the difference of the head and tail categories, we further develop a saliency-guided mixing strategy, Shap-Mix. Concretely, we first utilize the Shapley value to perform the saliency estimation for different body parts of each action category. Then utilizing the obtained saliency maps, a tail-aware mixing policy is proposed to maintain the representative motion patterns for the tail categories. Shap-Mix promotes decision boundary learning for the minority categories by explicitly establishing the relationship between crucial motion patterns and high-level semantics.
We conduct extensive experiments to verify the effectiveness of our method. Remarkably, a significant improvement is achieved with our method under both balanced and long-tailed settings.
 
Our contributions can be summarized as follows:
\begin{itemize}[leftmargin=1em]
\item We propose an effective skeleton mixing method, Shap-Mix, for long-tailed action recognition, which consists of a novel skeleton saliency estimation technique based on the Shapley value in cooperative game theory. It jointly considers the relationship between different skeleton joints to obtain the corresponding importance more rationally.

\item Based on the obtained joint importance distribution, a tail-aware mixing strategy is proposed, which prefers to produce the mixed samples that are more representative for the tail categories. It alleviates the over-fitting problem of tail categories by explicitly establishing the relationship between crucial motion patterns and high-level semantics.

\item A large-scale benchmark for long-tailed skeleton action recognition, covering three popular datasets and different methodological algorithms, is provided to benefit the community. Our method demonstrates the significant performance improvement for the long-tailed learning, and the notable effectiveness is also verified in the balanced setting with full datasets.
\end{itemize}

\section{Related Works}
\subsection{Skeleton-based Action Recognition}
Skeleton-based action recognition aims \whzjh{to} classify the action categories using 3D coordinates data of \whzjh{the} human body. 
Previous works are mostly based on the \whzjh{recurrent neural network (RNN)}, and \whzjh{convolutional neural network (CNN)}, treating the skeleton in the temporal series~\cite{song2017end,song2018skeleton,song2018spatio} or pseudo 2D-image~\cite{ke2017new,liu2017enhanced}.
Recently, inspired by the natural topology structure of \whzjh{the} human body, graph convolutional neural network (GCN)-based methods~\cite{yan2018spatial,shi2019two,cheng2020skeleton} have attracted more attention, achieving remarkable performance.
Meanwhile, \whzjh{transformer}-based models~\cite{shi2020decoupled,plizzari2021skeleton} also show promising results \whzjh{by learning long-range temporal dependencies}, owing to the attention mechanism. 

Different from the above works on model architecture, we focus on mixing augmentation design in this paper, which improves the performance as a plug-in design.

\subsection{Long-Tailed Visual Recognition}
There are two typical methods to tackle long-tailed learning, \ie, re-sampling and re-weighting methods. Re-sampling methods~\cite{van2007experimental,han2005borderline,shen2016relay} deal with the data imbalance issue by oversampling the tail categories or under-sampling the head categories. Re-weighting methods~\cite{lin2017focal,ren2020balanced,park2021influence}, often assign a higher weight to the tail categories to balance the positive gradients and negative gradients flowing. 
Recently, valuable efforts have been also made on the data augmentation~\cite{chou2020remix,du2023global}, ensemble learning~\cite{wang2020long}, decoupled learning~\cite{kang2019decoupling}, and contrastive learning~\cite{zhu2022balanced}, providing new perspectives on the image long-tailed learning.

In contrast, the long-tailed issue has not been well explored for skeleton data, especially for the modality-specific data augmentation methods. Previous works targeted at image data can be unsuitable and sub-optimal due to the sparse and compact body structure in the skeleton. Meanwhile, some skeleton augmentation methods~\cite{xu2022topology,chen2022skelemixclr} do not well consider the crucial spatial-temporal dynamics in human skeleton, especially under the long-tailed distribution, leading to the generation of less representative data samples. 
To this end, we propose a customized mixing augmentation with saliency guidance to generate new data samples to benefit the long-tailed learning.

\subsection{Shapley Value}
\label{sec:shap}
Shapley value~\cite{shapley1953value} is one of the most important concepts in cooperative game theory. It is used to allocate the achieved overall worth in cooperation to each player. Let us consider the set $\mathbb{P}$ and a function $f(\cdot)$ indicating the corresponding worth achieved by some players as a real number. The Shapley value of player $i$ is its average marginal contribution to all possible coalitions $\mathbb{S} \subseteq \mathbb{P}$ that can be formed without $i$. Concretely, it is formulated as
\begin{equation}
    \centering
    SV(i) = \frac{1}{|\mathbb{P}|} \sum_{\mathbb{S}\subseteq (\mathbb{P}-\{ i \}) } \frac{ f(\mathbb{S} \cup {i}) - f(\mathbb{S}) }{comb(|\mathbb{S}|, |(\mathbb{P}-\{i\})|)},
\end{equation}
where the $comb(\cdot,\cdot)$ is the function of combination number. The obtained $SV(i)$ is the final allocated 
worth to the player $i$. It well considers the interaction between different players, giving a more rational allocation solution. Meanwhile, it is shown that Shapley value satisfies good properties, \eg, \textit{Efficiency, Symmetry, Linearity} and \textit{Null player}~\cite{hart2016bibliography}. 

Due to its solid theory foundation and desirable properties, Shapley value has attracted much attention for machine learning study.%, \eg, feature selection~\cite{patel2021game} and explainability of deep learning~\cite{zheng2022shap,ghorbani2020neuron}. 
However, one drawback is its high computation cost as an NP-complete problem~\cite{deng1994complexity}. In response, many works adopt the Monte Carlo Sampling to give an estimation.

In this paper, we innovatively apply the Shapley value to the skeleton saliency estimation, which is the first to our best knowledge. Thanks to the sparsity of skeleton data, our approach does not require too much computational overhead.

\section{The Proposed Method: Shap-Mix}
We introduce our proposed method in this section. First, a simple yet effective skeleton mixing augmentation is presented. Based on this, we further propose a customized rebalanced mixing strategy guided by Shapley value to handle the long-tailed data distribution. Finally, the whole training scheme of the model is provided.

\subsection{Preliminaries}
\label{sec:pre}
\paragraph{Notations.} 
Generally, a skeleton sequence (the $i_{th}$ sample) can be represented as $s_i \in \mathbb{R}^{C\times T \times V \times M}$ with its label as $c_i$, where $C, T, V, M$ are channel, frame, joint, and performer dimensions, respectively. Note that the number of head categories is often much larger than the tail ones in long-tailed recognition. Our goal is to train a model with good generalization capacity on different categories using long-tailed data. %causing the over-fitting problem and the poor generalization capacity on tail classes. 

\paragraph{Mixup and Cut-Mix.} Mixup~\cite{zhang2017mixup} and Cut-Mix~\cite{yun2019cutmix} are proposed as a regularization technique to improve the generalization capacity of deep model. Two skeleton sequences, $s_i$ and $s_j$, are randomly sampled first. Then for Mixup, the two sequences are mixed in the spatial-temporal dimension $s^{Mix}_{ij} = \lambda * s_i + (1- \lambda) * s_j,$
where $\lambda$ is the mixing ratio number, sampled from the beta distribution. And for Cut-Mix, a spatial-temporal binary mask $m$ is generated, with a masking ratio of $1-\lambda$. Then the mixing operation is applied in a copy-paste manner, \ie, $s^{Mix}_{ij} = m \odot s_i + (1- m) \odot s_j,$
where $\odot$ is the element-wise multiplication operator.
The corresponding mixed label in the two methods is obtained by $c^{Mix}_{ij} = \lambda * c_i + (1- \lambda) * c_j$.

\subsection{A Simple Skeleton Mixing Strategy}
\label{sec:naive}

Mixing methods have been proven effective for representation learning and well-explored in image field~\cite{qin2020resizemix,walawalkar2020attentive,liu2022automix}. However, for skeleton data, there is still a lack of effective mixing design considering its complex spatial-temporal dependencies. To this end, we introduce a simple yet effective mixing method, ST-Mix, for skeleton data in the following.

\paragraph{Spatial-Temporal Mixing Design.}
To fully boost the modeling of skeleton spatial-temporal dynamics, we separately consider the mixing strategy on the spatial and temporal dimensions. Note that it is for Cut-Mix to generate a better mask, while is not required in Mixup.

For spatial dimension, many works~\cite{hua2023part,chen2022skelemixclr} have found that meaningful motion patterns are captured at the part level instead of the joint level. Therefore, we first divide the skeleton into five different parts, \ie, \textit{trunk, left arm, right arm, left leg, }and \textit{right leg} following the previous works. Then the mixing operation occurs at the level of body parts to maintain the crucial motion patterns. Specifically, we randomly sampled $N_s \in [N_s^l, N_s^u] $ body parts as the mixing targets of the spatial dimension. $N_s^l$ and $N_s^u$ are the lower- and upper-bound hyper-parameters.

For the temporal dimension, we can simply copy a sub-clip sampled from $s_j$ into the corresponding position of $s_i$ directly. However, a short clip can not accurately represent the whole action sequence for model to recognize. For example, when only a short clip of a hand in raising is observed, many actions, \eg, \textit{making phone calls} and \textit{drinking}, can be the candidates, leading to the confusion of labels. Therefore, we propose a temporal down-sampling strategy to construct more informative mixed samples. Specifically, supposing $N_t$ frames are to be mixed, a clip is sampled from $s_j$ with its length in $[N_t, T]$ and then down-sampled into $N_t$ frames to paste into $s_i$ to generate the final mixed sample. 

\paragraph{Joint Global-Local Mixture Learning.}
As discussed before, Mixup and Cut-Mix construct the mixed samples in different means, where the former produces the global mixture while the latter performs the copy-paste operation on the local patterns. To introduce more diverse motion patterns and further improve the generalization capacity, we bake these two mixture generation paradigms into our training process. Specifically, for each skeleton, we randomly choose and apply a mixing method, \ie, Mix-up or Cut-Mix, jointly learning the global and local mixture semantics. 

Here we highlight these specific points to distinguish our method from other skeleton mixing works~\cite{xu2022topology,chen2022skelemixclr}, 1) the well-designed mixing policy in spatial-temporal dimensions, 2) the joint learning for the global and local mixed data. These simple yet effective designs significantly improve the model performance compared with other mixing methods as shown in Table~\ref{tab:mix}, providing a strong skeleton augmentation technique.
Besides, different from other works, we focus on supervised learning, especially under the long-tailed data distribution, which will be discussed in the following sections.

\begin{figure}[t]
    \centering
    \includegraphics[width=0.49\textwidth]{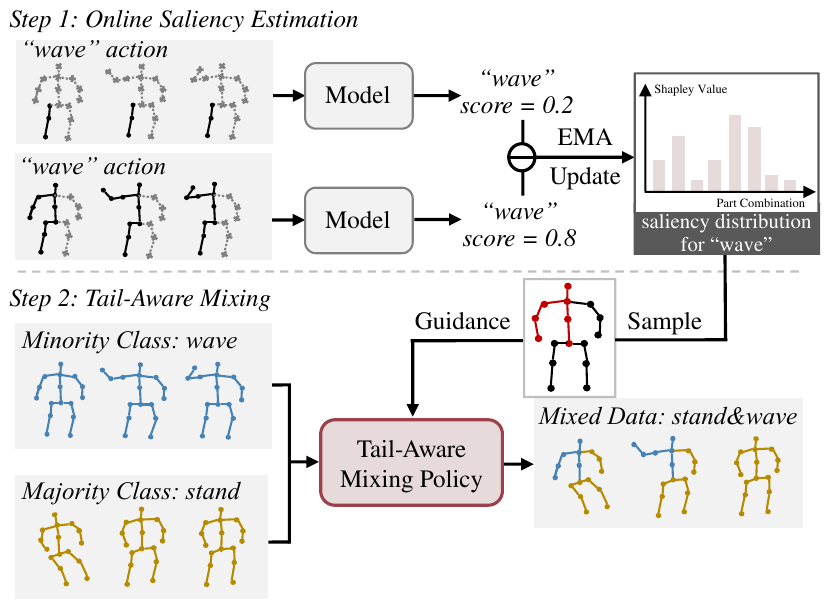}
   \caption{
   A simplified illustration of Shap-Mix. 
   We first perform the online saliency estimation using Eq. (2). In this example, $r =$\{\textit{right leg}\} and $b =$ \{\textit{trunk, right arm}\}. For dotted joints, we use the mean of the dataset as the static sequence. The calculated Shapley value is used to update the Shapley value list $v^c$ by EMA. Finally, the mixed data is generated, preserving the representative motion patterns of the minority class (\textit{wave} in this example).
   }
   \vspace{-5pt}
  \label{fig:pipeline}
\end{figure}

\subsection{Shapley-Value-Guided Rebalanced Mixing}
Now we have obtained an effective skeleton mixing augmentation. However, it adopts a random mixing manner, \ie, treating all categories equally during training. Considering the data distribution in real world is often long-tailed, it can be unreasonable because tail categories often necessitate more struggle to learn the underlying distribution. As a remedy to this issue, we further propose a customized mixing strategy, Shap-Mix, guided by Shapley value~\cite{shapley1953value} to promote the decision boundary learning for tail categories as shown in Figure~\ref{fig:pipeline}. Our key idea is to 1) obtain the saliency map of the skeleton joints first and then 2) use it to guide the synthesis of higher quality mixed samples for tail category learning. More details are provided as follows.

\paragraph{Part Saliency Estimation Based on Shapley-Value.} Shapley value, as introduced in Section~\ref{sec:shap}, is effective on the allocation of the worth to a set of players in cooperation. In addition to its own good properties and solid theoretical foundation, we list three key factors that motivate our choice of the Shapley value for skeleton saliency estimation: 

\noindent $\bullet$ Shapley value takes into account the interactions between different players, \ie, body parts in our context, to give the saliency. For example, hands (in raising) are important to recognize \textit{saluting}, while it can also be confused with other actions, \eg, \textit{making phone calls} and \textit{drinking} with similar hand motions. Therefore, the interaction between human trunk and hands should be considered jointly when measuring the importance of hands to action \textit{saluting}.

\noindent $\bullet$ Shapley value is based on the input instead of the deep feature maps. It can avoid the adverse effect of over-smoothing~\cite{liu2020towards} problem, the exponential convergence of similarity measures on node features, which is widely existing in GCNs. In other words, the saliency estimation can be inaccurate using the methods based on the fusion of deep features, \eg, Grad-CAM~\cite{selvaraju2017grad}.

\noindent $\bullet$ A disadvantage of Shapley value is that its calculation can be computationally costly. Fortunately, skeleton is much sparser compared with image data, which can effectively reduce the computation overhead. Besides, we adopt an online estimation method to further improve the efficiency.

Specifically, we maintain a saliency value list $v^c$ for each category $c$, which stores $v_b^c$ for every body part combination $b$, \ie, $v^c = \{v_b^c\}$. Specifically, $b$ denotes a body part combination that may appear in the spatial mixing. It is a subset of the universal set $\mathbb{U}$ containing all pre-defined $5$ body parts, \eg, $b=$\{\textit{left arm}, \textit{right leg}\}. 
The overall worth in our context is defined as the predicted confidence of the model on the correct class $c$, \ie, $f^c (\cdot)$. Then the Shapley value $v_b^c$ is computed as:
\begin{equation}
    \centering
    \label{eq:shap_skeleton}
    v_b^c = \frac{1}{|\mathbb{U}-b|+1}\sum_{r\subseteq (\mathbb{U} - b)}\frac{f^c(r\cup b) - f^c(r)}{comb(|r|,|(\mathbb{U} - b)|)},
\end{equation}
where $comb(\cdot,\cdot)$ returns the combination number and $|\cdot|$ is the cardinality of a set. Intuitively, $v_b^c$ represents the average
marginal contribution of part combination $b$ to the prediction confidence on the class $c$ it belongs to. 

Next, we provide some more implementation details for a better understanding. First, $v_b^c$ is computed and averaged only on the skeletons belonging to class $c$, and all these skeletons share the same saliency list $v^c$. When computing the prediction score in Eq.~(\ref{eq:shap_skeleton}), the selected body parts, together with the other part regions which are set to the corresponding mean joint value of the dataset, make up the complete input skeleton for model. For efficiency, we do not consider the Shapley value of temporal dimension, \ie, we always replace the same body parts in all the frames for saliency estimation. Meanwhile, we compute the Shapley value in an online way during training to share the computational overhead. Specifically, we only sample the part combination $r$ and $b$ once to calculate the single marginal contribution at each iteration, and then update the $v_b^c$ using exponential moving average (EMA) to get the average estimation during the whole training process. The obtained saliency results will be utilized to guide the mixing synthesis as introduced in the following.
\paragraph{Tail-Aware Mixing Synthesis.}
We find that the intra-class saliency estimation is easier than the inter-class decision boundary learning in Figure~\ref{fig:saliency}. Therefore, the obtained saliency maps can serve as guidance for the synthesis of mixed data. Due to the scarcity of tail class samples, we suggest explicitly guiding the model to capture the relationships between the crucial motion patterns and the action semantics, revealing the possible underlying distributions of tail categories. Therefore, we introduce a tail-aware mixing policy, to produce more representative mixed data for tail classes.

Specifically, given the source data $s_i$ with label $c_i$, and the target data $s_j$ with label $c_j$, the body parts to be mixed are sampled from a specific importance 
distribution $d(\cdot)$, which can be formulated as:
\begin{equation}
    \centering
    d(c) = softmax(norm(\{v^c_b/|b|\})/\tau),
\end{equation}
where $norm(\cdot)$ is the $l_1$ normalization and $|b|$ is the joint number in the part combination $b$. $\tau$ is the temperature hyper-parameter. A bigger $v_b^c/|b|$ indicates a higher probability that the part combination $b$ will be selected.
Specifically, two possible cases are discussed: 

\noindent$\bullet$ If the sample number of class $c_i$ is more than $c_j$, a random body part combination would be sampled from $d(c_j)$, which is more likely representative for class $c_j$. Then the selected body parts of $s_j$ are pasted into $s_i$ to generate the mixed data. 

\noindent$\bullet$ If the first case is not true, a body part combination would be sampled from $d(c_i)$. Then the actual body parts to be mixed are the complement of the sampled parts in $s_j$. They are pasted into $s_i$ to preserve the crucial motion patterns in minority categories.

By virtue of this, the mixed sample is expected to be representative for the minority categories. Although the majority categories are not explicitly considered in the mixing synthesis, we assume that the model can learn desirable representations for them using sufficient training samples in datasets. As for the label of mixed data, we simply follow the same definition as discussed in Section~\ref{sec:pre} to make our method more general and compatible with other re-weighting methods.

\subsection{Training with Shap-Mix}
In \textit{long-tailed} recognition task, we utilize the \textit{Cross-Entropy} loss with \textit{Balanced Softmax}~\cite{ren2020balanced} to tackle the data imbalance problem as discussed in Section~\ref{sec:intro}. During training, the proposed Shap-Mix technique is integrated, and the model optimizes the loss objective for the mixed data and original data jointly. Note that the saliency value obtained by the model can be inaccurate at the beginning of training. Therefore, we warm up the model in the first few epochs to obtain a stable estimation of Shapley-value, after which we use the estimation to guide the mixing data generation.
Meanwhile, we compare our method with previous works in \textit{balanced} setting to verify the effectiveness. Specifically, we utilize the proposed ST-Mix in Section~\ref{sec:naive} for efficiency because of the balanced class distributions in this setting.

\section{Experiment Results}
\subsection{Datasets}
\paragraph{NTU RGB+D 60 Dataset (NTU 60)}~\cite{shahroudy2016ntu}. There are 56,578 videos with skeleton data of 25 joints, divided into 60 action categories.
Two evaluation protocols are recommended:  a) Cross-Subject (xsub): the training data are collected from 20 subjects, while the testing data are from the other 20 subjects. b) Cross-View (xview):  the front and two side captured views are used for training, while testing set includes the left and right 45-degree views. 

\paragraph{NTU RGB+D 120 Dataset (NTU 120)}~\cite{liu2019ntu}. This is an extension to NTU 60, consisting of 114,480 videos with 120 categories. Two recommended protocols are presented: a) Cross-Subject (xsub): the training data are collected from 53 subjects, while the other 53 subjects are for testing. b) Cross-Setup (xset): the training data use even setup IDs, while testing data use odd ones.

\paragraph{Kinetics Skeleton 400 (Kinetics 400)}~\cite{kay2017kinetics}.
This dataset contains the 2D skeleton data for action recognition, extracted from the Kinetics 400 video dataset using \textit{OpenPose} toolbox. It is the largest skeleton-based action recognition dataset, containing more than 260k training sequences over 400 classes in a long-tailed distribution. The sample number for each category ranges from 200 to 1000.

For the imbalanced setting, we construct the long-tailed datasets based on NTU 60 and 120 (\textbf{LT-NTU 60/120}), and adopt the cross-subject evaluation protocol following~\cite{liu2023balanced}. The \textit{imbalanced factor} (IF) is defined as the number of
training samples in the largest class divided by the smallest~\cite{cui2019class}. Meanwhile, Kinetics 400 is used as a real long-tailed benchmark in the wild.

\subsection{Implementation Details}
Our method can be applied to any backbone for skeleton-based action recognition. Specifically, CTR-GCN~\cite{chen2021channel} is chosen as our backbone for comparison and we follow its training settings and data processing for fairness. %The results of different backbones can be found in \textit{Supplementary Material}.
For the implementation of Shap-Mix, we randomly sample 2 or 3 body parts to mix in spatial dimension. The mixed temporal length is from 40\% to 70\% of the original length. The temperature $\tau$ is set as 0.2, and the momentum coefficient in the EMA is 0.9. The warm-up phase is for the first 5 epochs. Due to the less data in the constructed long-tailed dataset, we increase the training epochs to 100.

\subsection{Comparison on Long-Tailed Recognition}
\begin{table*}[t]
\small
\centering
  
\resizebox{\linewidth}{!}{%
  \begin{tabular}{l|c|c c c c| c c c c}
    \toprule
     \multirow{2}{*}{ {Method} } & \multirow{2}{*}{Category}& \multicolumn{4}{c}{ {LT-NTU  60 xsub (IF = 100)} } & \multicolumn{4}{c}{ {LT-NTU 120 xsub (IF = 100)} } \\
     		& & Overall & Many & Medium & Few & Overall & Many & Medium & Few  \\
    \midrule
    Cross-Entropy Loss & - & 74.4 &86.4 & 69.5& 63.8 &64.2& 83.6 & 64.9& 54.7\\
    \midrule
    Mixup~\cite{zhang2017mixup} & \multirow{5}{*}{ Augmentation} & 75.9 & 86.3 & 71.7& 66.5 &66.9 &  \textbf{85.6} &65.6&55.5\\
    Remix-DRW~\cite{chou2020remix} & &78.7 &86.6 &74.8 &72.7 & 69.3 & 83.5 & 67.2& 62.6 \\
    FSA~\cite{chu2020feature} & & 76.8 & 85.4 &73.3 &68.8 & 66.7 & 82.4 & 66.8 & 59.4 \\
    GLMC~\cite{du2023global} & & \underline{78.8} &78.6 & \textbf{81.3} & \textbf{76.0}&	\underline{71.5} & 79.5 & \underline{70.5}& \underline{69.2}\\
    BRL~\cite{liu2023balanced} & & 76.9 &85.2 &73.1 &70.0 & 66.3 &84.2 & 65.0 & 59.9 \\
    \midrule
    ROS~\cite{van2007experimental} &\multirow{4}{*}{Reweighting} & 74.8 &86.1 & 68.1& 57.9& 61.0 & 81.3& 62.3& 50.7\\
    Focal Loss~\cite{lin2017focal} &\multirow{4}{*}{\&}  & 77.6 &83.1&75.9&72.0 & 69.4 &81.7 & 66.7 &65.2\\
    CB Loss~\cite{cui2019class} &\multirow{4}{*}{Resampling} & 72.4 & 78.4& 71.2& 65.3& 63.2 &76.0& 65.0& 56.1\\
    LDAM-DRW~\cite{cao2019learning} & & 76.4 & 83.7 & 73.4 &69.9 & 66.8 & 80.9 & 65.8 & 61.2 \\
    Balanced Softmax (BS)~\cite{ren2020balanced} & & 77.6 & 83.8 & 73.9 & 73.3 &69.6 & 81.4& 67.8& 66.7 \\
    IB Loss~\cite{park2021influence} & & 76.0  &84.0&74.4& 66.5& 67.8 &81.2& 67.6& 62.4\\
    \midrule
    BS+Max Norm~\cite{LTRweightbalancing} & Decoupling & 77.5 & 81.2 & 75.9 & 74.1 &	70.3 & 80.0 & 68.2 & 68.8\\
    \midrule
    PaCo~\cite{cui2021parametric} & Contrastive & 76.6 
 & 82.0 & 76.4 & 69.1 & 67.9 & 82.2 & 66.2 & 63.8 \\
    BCL~\cite{zhu2022balanced} & Learning & 77.3 & 84.4 & 74.1 & 71.5& 66.9 & 82.3 & 64.5 & 62.8\\
    \midrule
    RIDE (3 experts)~\cite{wang2020long} & Ensemble & 76.6 & \underline{86.7} & 71.9 & 68.4 & 65.2 & \underline{85.4} & 66.7 & 53.8\\
    \midrule
    Shap-Mix (Ours) & \multirow{1}{*}{Augmentation} & \textbf{80.8} & \textbf{86.8} & \underline{78.4} & \underline{75.6} & \textbf{73.0 }&{84.8} & \textbf{71.3} & \textbf{69.7}\\
    \bottomrule
\end{tabular}
}
\caption{Performance comparison of long-tailed skeleton-based action recognition with single joint stream. IF is the imbalance factor. Top-1 accuracy (\%) is reported. The results with \textbf{bold} and \underline{underline} indicate the highest and second-highest value.}

  \label{tab:imbalance}
\end{table*}

We compare the popular long-tailed recognition works including different methodological categories. Specifically, decoupling-based methods utilize two-stage training to decouple the learning of the feature extractor and classifier. Contrastive-learning-based methods apply balanced contrastive representation learning. Ensemble-based methods employ multiple experts and perform the knowledge ensemble to obtain the final predictions. Since these methods are not implemented on the skeleton data, we conduct extensive reproductive experiments to provide a benchmark. \textit{We strictly follow the setting fairness including training epochs and utilize the same backbone}, using the official implementation code as possible.

\begin{table}[t]
\centering
  \small
  \resizebox{\linewidth}{!}{%
  \begin{tabular}{l|c c c c c}
  \toprule
  Method & Baseline & PoseConv3D & BRL$^\dag$ & GLMC$^\dag$ &Ours$^\dag$ \\
  \midrule
  Acc. (\%) & 45.0 & 46.0 & 45.6 & 46.6 & \textbf{48.4}\\
  \bottomrule
\end{tabular}
}\caption{Comparison results on Kinetics 400 of single joint stream. $^\dag$ indicates the long-tailed methods.}

  \label{tab:k400}
\end{table}

\begin{table}[t]
\centering
  \small
  \resizebox{0.93\linewidth}{!}{%
  \begin{tabular}{l|c c c}
  \toprule
  Method & IF = 10 (\%)& IF =50 (\%) & IF = 100 (\%)\\
  \midrule
  Baseline & 79.6 & 70.1 & 64.2\\
  Remix-DRW & 81.6 & 74.4 &69.3\\
  GLMC & 82.0 & 75.9 & 71.5 \\
  Balanced Softmax &80.7 & 74.2 &69.6\\
  Shap-Mix (Ours) & \textbf{83.0} & \textbf{76.8} & \textbf{73.0}\\
  \bottomrule
\end{tabular}
}\caption{The results under LT-NTU 120 dataset of different imbalance factors (IF).}

  \label{tab:imb}
\end{table}

Following~\cite{liu2023balanced,du2023global}, we first report the accuracy on LT-NTU datasets of three splits of classes, Many-shot classes
(training samples$>$100), Medium-shot (training samples 20$\sim$100) and Few-shot (training samples$<$20), to comprehensively evaluate our model. As we can see in Table~\ref{tab:imbalance}, our method achieves the best overall scores compared with different competitors. Notably, compared with GLMC, which is the SOTA long-tailed augmentation method, our method shows a desirable overall performance improvement, especially for the head categories. Meanwhile, compared with the baseline, our method can largely boost the performance on the medium- and few-shot classes without compromising that on many-shot classes, which is difficult for most long-tailed methods.

We also conduct the experiments on Kinetics 400, which is a long-tailed dataset in the wild, as shown in Table~\ref{tab:k400}. Compared with PoseConv3D~\cite{duan2022revisiting} and GLMC, which are the SOTAs in standard supervised and long-tailed learning methods, Shap-Mix achieves the best scores owing to the utilization of the crucial motion patterns.

Finally, we present the results with different imbalance factors in Table~\ref{tab:imb}. Our method can improve both general representation quality and decision boundary learning for tail categories, achieving the best scores across different imbalance factors compared with other long-tailed methods.

\subsection{Comparison on Balanced Recognition}

\begin{table}[t]
\small
\centering
\resizebox{0.95\linewidth}{!}{%
  \begin{tabular}{l|c| c c c c}
    \toprule
     \multirow{2}{*}{ {Method} } & \multirow{2}{*}{ {Year} } & \multicolumn{2}{c}{ {NTU  60 (\%)} } & \multicolumn{2}{c}{ {NTU 120 (\%)} } \\
     		& & {xsub} & {xview} & {xsub} & {xset} \\
    \midrule
    2s-SGN & CVPR'20 & 89.0 & 94.5 & 79.2 & 81.5 \\
    4s-Shift-GCN & CVPR'20 & 90.7 & 96.5 & 85.9 & 87.6 \\
    2s-MS-G3D & CVPR'20 & 91.5 & 96.2 & 86.9 & 88.4 \\
    4s-MST-GCN  & AAAI'21 & 91.5 & 96.6 & 87.5 & 88.8 \\
    4s-CTR-GCN  & ICCV'21 & 92.4 & 96.8 & {88.9} & {90.6} \\
    4s-Info-GCN & CVPR'22 &92.7 &96.9 &89.4 &90.7 \\
    6s-Info-GCN & CVPR'22 &93.0 &\textbf{97.1} &89.8 &91.2 \\
    3s-EfficientGCN & TPAMI'22 & 91.7 & 95.7 & 88.3 & 89.1 \\
    4s-FR-Head & CVPR'23 & 92.8 & 96.8 & 89.5 &90.9 \\
    6s-StreamGCN & IJCAI'23 & 92.9 & 96.9 &89.7 & 91.0\\
    4s-HD-GCN &ICCV'23 &93.0& 97.0 &89.8 &91.2\\
    4s-STC-Net &ICCV'23 &93.0 &\textbf{97.1} &89.9 &91.3\\
    \midrule
    2s-Ours & -& 93.4 & 96.8 & 90.2 & 91.6 \\
    4s-\textbf{Ours} & - & \textbf{93.7} & \textbf{97.1} & \textbf{90.4} &  \textbf{91.7} \\
    \bottomrule
\end{tabular}
}
\caption{Performance comparison of balanced recognition on NTU datasets in top-1 accuracy. $\ast$s- means the fusion results of $\ast$ streams.}

  \label{tab:balance}
\end{table}

Here we show the effectiveness of our method under balanced setting. The model is trained on the balanced (original) NTU datasets to give a fair comparison with previous skeleton-based action recognition methods.
We first compare our method with the state-of-the-art methods, including 
SGN~\cite{zhang2020semantics}, 
Shift-GCN~\cite{cheng2020skeleton}, MS-G3D~\cite{liu2020disentangling}, MST-GCN~\cite{chen2021multi}, CTR-GCN~\cite{chen2021channel}, Info-GCN~\cite{chi2022infogcn}, EfficientGCN~\cite{song2022constructing}, FR-Head~\cite{zhou2023learning}, StreamGCN~\cite{yang2023action}, HD-GCN~\cite{lee2023hierarchically}, STC-Net~\cite{lee2023leveraging}. Following previous works, we report the results of multi-stream fusion, \ie, joint, bone (2-stream), joint motion and bone motion (4-stream). As shown in Table~\ref{tab:balance}, our method achieves the best performance across different datasets. \textbf{Remarkably}, our method with 4 (/2) streams can outperform many latest methods with 6 (/4) streams, verifying the significant effectiveness.

Meanwhile, our augmentation method can be equipped with different backbones. Compared with another model-agnostic method, FR-Head~\cite{zhou2023learning}, our method can bring consistent performance improvement without introducing additional training parameters as shown in Table~\ref{tab:backbones}. %Compared with another model-agnostic method, FR-Head~\cite{zhou2023learning}, our method show desirable advantages in both performance and parameters.

\begin{table}[t]
\centering
  \small
  \begin{tabular}{l c c c}
  \toprule
  { {Method} } & {{Params.}} & {xsub (\%)} & {xset (\%)} \\
  \midrule
  2s-AGCN & 3.80M & 84.3 & 85.9 \\
  ~~+ FR Head & 4.33M & 84.6$^{\uparrow 0.3}$ & \textbf{86.6}$^{\uparrow 0.7}$ \\
  ~~+ Ours & 3.80M & 84.6$^{\uparrow 0.3}$ & 86.5$^{\uparrow 0.6}$ \\
  \midrule
  CTR-GCN & 1.46M & 84.5 & 86.6 \\
  ~~+ FR Head & 1.99M & 85.5$^{\uparrow 1.0}$ & 87.3$^{\uparrow 0.7}$ \\
  ~~+ Ours & 1.46M & \textbf{86.9}$^{\uparrow 2.4}$ & \textbf{88.3}$^{\uparrow 1.7}$\\
  \midrule
  Info-GCN & 1.58M & 85.1 & 86.3 \\
  ~~+ FR Head & 2.11M & - & - \\
  ~~+ Ours & 1.58M & \textbf{85.7}$^{\uparrow 0.6}$ & \textbf{88.0}$^{\uparrow 1.7}$\\
  \midrule
  HD-GCN  & 1.68M & 85.1 & 87.2 \\
  ~~+ FR Head & 2.21M & 85.4$^{\uparrow 0.3}$ & 87.7$^{\uparrow 0.5}$ \\
  ~~+ Ours & 1.68M & \textbf{87.0}$^{\uparrow 1.9}$ & \textbf{88.8}$^{\uparrow 1.6}$\\
  \bottomrule
\end{tabular}
\caption{Performance of our proposed method using different backbones with single joint  stream on NTU 120 dataset. The results of previous work, FR-Head, are given for comparison.}

  \label{tab:backbones}
\end{table}

\subsection{Ablation Studies}
Next we present the ablation results conducted on the (LT) NTU 60 dataset under cross-subject protocol.

\paragraph{Effectiveness of ST-Mix Design.} We compare our ST-Mix with different mixing methods for skeleton data in Table~\ref{tab:mix}. The reported results are obtained under the balanced action recognition with the full NTU 60 dataset. Compared with Mixup and Cut-Mix, our method jointly learns these two mixture patterns, yielding a notable performance improvement. Meanwhile, due to the well-designed spatial-temporal mixing policy, our method outperforms other mixing methods, with their respective effects presented in Table~\ref{tab:st_design}.

\begin{table}[t]
            \vspace{0pt}
		\parbox{.52\linewidth}{
            \vspace{0pt}
  \small
  \centering
  \begin{tabular}{l| c}
  \toprule
  { {Method} }  & {xsub}\\
  \midrule
  Baseline & 89.9\\
  \midrule
   Mixup & 90.8\\
   Cut-Mix & 90.7\\
   Mix~\cite{xu2022topology} & 90.1\\
   Mix~\cite{chen2022skelemixclr} & 90.7 \\
  \midrule
   ST-Mix (Ours) & \textbf{91.6}\\
  \bottomrule
\end{tabular}
\caption{Comparison with other mixing methods. The latter two are also designed for skeleton.}
\label{tab:mix}
		}\hfill
            \vspace{0pt}
		\parbox{.44\linewidth}{
			\centering
                \small
			\setlength{\tabcolsep}{6pt}
			\begin{tabular}{c c|c}
                   \toprule
                    Spa. & Temp. & xsub \\
                    \midrule
                     \checkmark & & 91.1\\
                     & \checkmark & 90.1\\
                     \checkmark &\checkmark &\textbf{91.6}\\ 
                    \bottomrule
                \end{tabular}
                \caption{Ablation study on spatial-temporal mixing design in ST-Mix. ``Spatial" is mixing in the part-level or joint-level. ``Temporal" denotes to copy directly or after randomly down-sampling.}
                  \label{tab:st_design}
		}\hfill
	\end{table}	

\begin{figure}[t]
    \centering
    \includegraphics[width=0.44\textwidth]{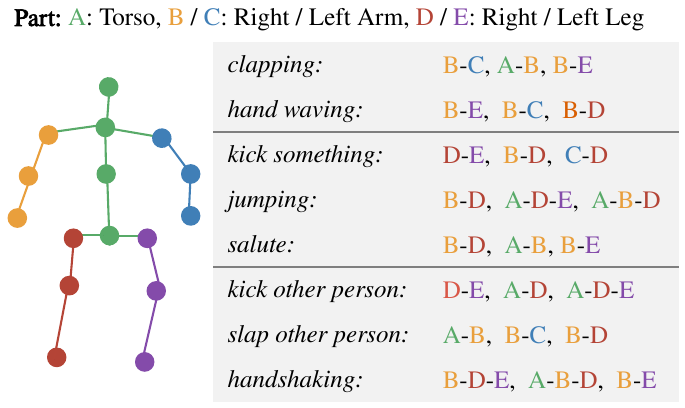}
   \caption{
   The visualization of our saliency estimation, in the form of \textit{action}, the first, second, and third most salient part combination. We choose the actions from many- (first 2), medium- (3-5), and few-shot (last 3) classes.}
   \vspace{-5pt}
  \label{fig:saliency}
\end{figure}

\paragraph{Visualization of Saliency Estimation.} %We show some saliency estimation results based on Shapley value in Figure~\ref{fig:saliency}. 
We choose the part combination containing 2 or 3 parts and estimate the corresponding importance. As shown in Figure~\ref{fig:saliency}, we can obtain a rational skeleton saliency estimation for many-, medium-, and few-shot classes. For example, in \textit{handshaking} action, the most salient part combination is obtained as the two legs with the right arm, which is reasonable because people usually stand with their hands out in this action. Meanwhile, it can be found hands and arms play an important role in most human actions, especially the dominant (right) hand of most people. 
These results are promising for more fine-grained action recognition and spatial localization. We hope that more works will emerge in the future to utilize and explore skeletal saliency maps.

\paragraph{Effectiveness of Shapley Value Guidance.}
The results of different guidance for mixing augmentation are provided in Table~\ref{tab:guidance}. As we can see, the model outperforms the mixing method without guidance either using re-samping or re-weighting method. Meanwhile, we compare with another guidance, Grad-CAM~\cite{selvaraju2017grad}. However, due to the over-smoothing problem, it can not achieve desirable performance improvement. In contrast, we use Shapley value as guidance, which is based on input and well considers the relationship between different joints, and brings further performance improvement. %The further complexity analysis of the method can be found in \textit{Supplementary Material}.

\begin{table}[t]
\centering
  \small
  \begin{tabular}{l c c}
  \toprule
  Mixing Guidance & RS-Acc. (\%) & RW-Acc. (\%) \\
  \midrule
  Random &  76.3 &80.0 \\
  Grad-CAM &  76.4 & 79.0 \\
  Shapley value &\textbf{77.9} & \textbf{80.8} \\
  \bottomrule
\end{tabular}
  \caption{Ablation study on the mixing with different guidance. The re-sampling (RS-) and re-weighting (RW-) techniques are applied to show the effect, respectively.}
  \label{tab:guidance}
\end{table}

\section{Limitation and Conclusion}
In this paper, we explore the long-tailed skeleton-based action recognition, which has been largely overlooked before, and propose a novel augmentation-based method, Shap-Mix. Specifically, we develop a saliency estimation method based on Shapley value and a tail-aware mixing policy to preserve more representative motion patterns, improving decision boundary learning of tail classes. One limitation is that due to the inherent computational complexity of Shapley value, our method is primarily for sparse data such as skeleton. This can be alleviated by performing estimation every $n$ iterations, which is a trade-off between estimation accuracy and computation overhead. Extensive experiments on both balanced and long-tailed settings verify the effectiveness.

\section*{Acknowledgments}
This work was supported in part by the National Natural Science Foundation of China under Grant 62172020, and in part by the State Key Laboratory of Media Audio and Video (Communication University of China), Ministry of Education, China.

\small
\bibliographystyle{named}
\bibliography{ijcai24}

\begin{algorithm*}[ht]
	\caption{Learning algorithm of the proposed Shap-Mix}
	\label{alg:algorithm}
	\textbf{Input}: A batch $\mathbb{B}=\{(x_i,y_i)\}_{i=1}^l$ from the dataset\\
	\textbf{Parameter}: Network $f(\cdot)$; The maintained list of Shapley value $svList \in \mathbb{R}^{C*N}$, $C$ is the class number and $N$ is the number of different body part combinations; $T_{max}$ is the maximum training iteration; $p$ is the probability for Mixup; $perm$ is a randomly generated permutation; $warm\_iters$ is the iteration number for warm-up. $\alpha$ is the momentum coefficient in EMA.\\
 \textbf{Begin}
	\begin{algorithmic}[1] %[1] enables line numbers
		\STATE \textbf{for} $T=1, 2, 3, ..., T_{max}$ \textbf{do} 
		\\ \quad// \textit{Online sailiency estimation.}
            \STATE \quad Randomly sample the body parts $r_i$ and $b_i$ (defined in Eq. (2)) for each skeleton $x_i$;
            \STATE \quad Calculate the Shapley value $v_b^{c_i}$ by Eq. (2) in the main paper;
		\STATE \quad \textbf{for} $i=1, 2, 3, ..., l$ \textbf{do} 
            \STATE 	\quad\quad $svList[c_i, b_i] = svList[c_i, b_i]*\alpha + v_{b_i}^{c_i}*(1-\alpha)$;
            \STATE \quad \textbf{end for}
    \\ \quad// \textit{Construct the mixed data.}
            %\STATE \quad Generate a random permutation $perm$.
            \STATE \quad \textbf{for} $i=1, 2, 3, ..., l$ \textbf{do} // \textit{train for a batch}
		\STATE \quad\quad \textbf{if} $random.random() < p$ \textbf{then} // \textit{Perform Mixup}
            \STATE \quad\quad\quad Randomly sample $\lambda$ from a uniform distribution;
            \STATE \quad\quad\quad $x_{mix, i} = \lambda * x_i + (1-\lambda) * x_{perm[i]}$;
            \STATE \quad\quad\quad $y_{mix, i} = \lambda * y_i + (1-\lambda) * y_{perm[i]}$;
            \STATE \quad\quad \textbf{else} // \textit{Perform Cut-Mix}
            \STATE \quad\quad\quad Sample the body parts from the distribution $svList$ to obtain the binary mask $m$ according to Section 3.3 (2);
		\STATE \quad\quad\quad $x_{mix, i} = m \odot x_i + (1-m) \odot x_{perm[i]}$;
            \STATE \quad\quad\quad $y_{mix, i} = m.mean() * y_i + (1-m.mean()) * y_{perm[i]}$;
            \STATE \quad\quad \textbf{end if}
        \STATE \quad\textbf{end for}
	\\ \quad// \textit{A step of training.}
		\STATE \quad $\hat{y} = f(x)$;
            \STATE \quad \textbf{if} $T > warm\_iters$ \textbf{then}
            \STATE \quad\quad $\hat{y_{mix}} = f(x_{mix})$;
		\STATE \quad\quad Calculate the rebalanced loss $\pounds$ for $(x, y)$ and $(x_{mix}, y_{mix})$; 
            \STATE \quad \textbf{else}
            \STATE \quad\quad Calculate the rebalanced loss $\pounds$ for $(x, y)$;
            \STATE \quad \textbf{end if}
            \STATE \quad  Update model parameters by minimizing $\pounds$;
		\STATE \textbf{end for}
	\end{algorithmic}
 \textbf{End}
\end{algorithm*}

\newpage
\appendix
\section{Implementation Details}

\subsection{The Proposed Shap-Mix Algorithm}
We present an illustration example using our proposed framework for a better understanding. Our framework consists of two steps. First, in each iteration, we perform the online saliency estimation for the skeleton sequence, \eg, ``wave” action in our example. Then the Shapley value is calulated as the average
marginal contribution of part combination $b$ to the prediction confidence on its class. Concretely, we utilize the EMA to update the Shapley value list of each category during training. Next, we generate the mixed data to train the model. Specifically, we first random sample two skeletons, \eg, ``wave" and ``stand". Supposing the ``wave" is the relatively minority class, the body parts to be mixed are sampled from the distribution derived from the saliency distribution for the minority category, \ie, ``wave". The mixed sequence is expected to preserve the salient parts of the minority categories to be representative. The overall training algorithm is shown in Algorithm~\ref{alg:algorithm}

\subsection{Data Pre-processing}
For data pre-processing, we follow the CTR-GCN~\cite{chen2021channel}. Specifically, the skeleton sequences are all cropped into 300 frames. Then we utilize the \textit{Temporal Resize-Crop} augmentation to sample the sequence with 64 frames as input for NTU datasets. \textit{Random Rotation} is selected as the spatial augmentation for all method implementation. Notably, we sample the skeleton with 100 frames for Kinetics 400, following the previous practice~\cite{duan2022pyskl}.

For the construction of the long-tailed datasaets, \ie, LT-NTU 60 / 120, we follow the previous works~\cite{liu2023balanced,buda2018systematic}, which truncates a subset with the Pareto distribution from the balanced version while keeping the validation set unchanged. Specifically, the maximum sample number in each class is set as 600 in LT-NTU 60 / 120 (IF=100).

\begin{table*}[t]
  \small
  \centering
  
  \caption{Long-tailed action recognition with different model backbones on LT-NTU 60.}
  \begin{tabular}{l|c|c|c}
   \toprule
   Backbones & Baseline &\textit{w} Shap-Mix & Performance $\Delta$ \\
    \midrule
    CTR-GCN~\cite{chen2021channel} & 74.4 & \textbf{80.8} & 6.4\\
    ST-GCN++~\cite{duan2022pyskl} & 72.5 & \textbf{78.7} & 6.2\\
    Info-GCN~\cite{chi2022infogcn} & 71.2 & \textbf{77.5} & 6.3\\
    HD-GCN~\cite{lee2023hierarchically} & 73.4 & \textbf{79.8} & 6.4\\
    \bottomrule
\end{tabular}
  \label{tab:backbones}
\end{table*}

\begin{table}
		%\centering
            \vspace{0pt}
		\parbox{.45\linewidth}{
            \vspace{0pt}
		\caption{Spatial configuration study of ST-Mix. The number of mixed body parts is sampled from $[N_s^l, N_s^u]$.}
  \small
  \label{tab:spatial}
  \centering
  \begin{tabular}{ l l | c}
  \toprule
  {$N_s^l$} & $N_s^u$  & Accuracy (\%) \\
  \midrule
  1 & 2 & 91.0\\
  1 & 3 & 91.3\\
  2 & 3 & \textbf{91.6}\\
  2 & 4 & 91.5\\
  3 & 4 & 91.2\\
  \bottomrule
\end{tabular}
		}\hfill
            \vspace{0pt}
		\parbox{.45\linewidth}{
			\caption{Temporal configuration study of ST-Mix. The ratio of mixed temporal length is sampled from $[N_t^l, N_t^u]$.}
			\centering
                \small
			\setlength{\tabcolsep}{6pt}
                \label{tab:temporal}
			\begin{tabular}{l l |c}
                   \toprule
                    $N_t^l$ & $N_t^u$& Accuracy (\%) \\
                    \midrule
                    0.3 & 0.6 & 91.2\\
                    0.3 & 0.7 & 91.4 \\
                     0.4 & 0.7 & \textbf{91.6}\\
                     0.4 & 0.8 & 91.0\\
                     0.5 & 0.8 & 91.2\\
                    \bottomrule
                \end{tabular}
		}\hfill
	\end{table}	

\subsection{Training Settings}
Our method is implemented by PyTorch and trained with the SGD optimizer with a momentum of 0.9. All training processes are on 2 NVIDIA 2080 Ti GPUs with an initial learning rate of 1e-4. The batch size is set as 64. 

For the balanced training, the model is trained for 65 epochs in total. The learning rate is decayed by 0.1 at the 35$_{th}$ and 55$_{th}$ epoch. For long-tailed training, the training epoch number is increased to 100 due to the few training data. And the learning rate is decayed by 0.1 at the 60$_{th}$ and 80$_{th}$ epoch for NTU datasets. Differently, for the Kinetics 400 dataset, we utilize a cosine learning rate decay strategy to maintain the setting fairness with previous works~\cite{liu2023balanced,duan2022revisiting}.

\section{More Experimental Results}
\subsection{Class-Wise Comparison with Baseline Model}
\begin{figure*}[t]
    \centering
    \includegraphics[width=0.85\textwidth]{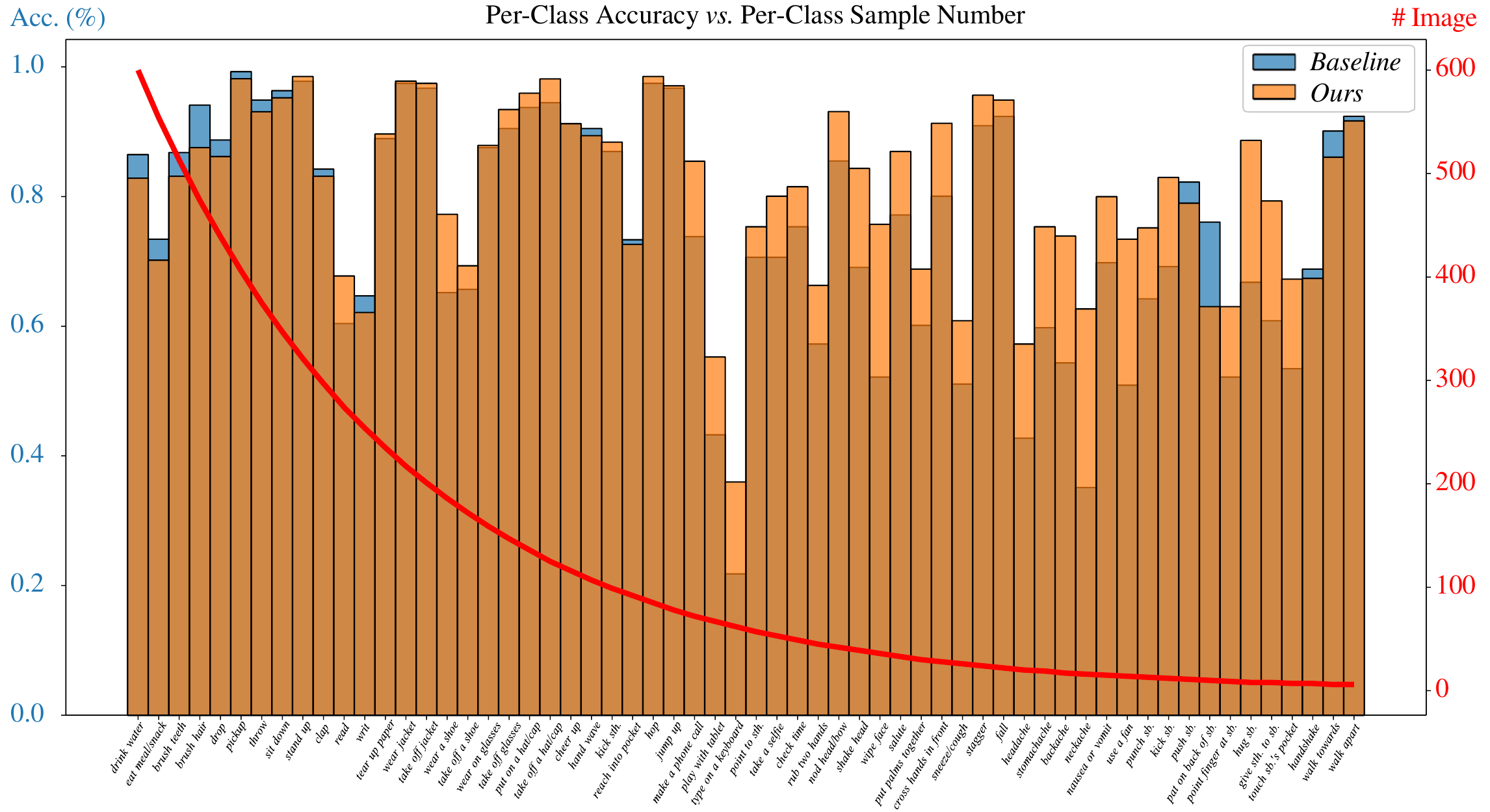}
   \caption{
   Comparison of our method with the baseline in terms of the accuracy of per-class. The sample number of each class is also presented using the red line.}
  \label{fig:class}
\end{figure*}

The class-wise results under long-tailed skeleton-based action recognition are provided in Figure~\ref{fig:class}. As we can see, our method can bring notable improvement for the minority categories compared with the baseline, while maintaining a decent performance for the majority categories. Although the salient parts for the majority categories are not considered in mixing, we argue that the abundant data samples can already imply a well-defined decision boundary for majority categories to obtain a good performance.

\begin{table}[]
  \small
    \centering
			\caption{Ablation study of temperature on LT-NTU 60 dataset.}
			\centering
                \small
			\setlength{\tabcolsep}{7mm}{
                \label{tab:temper}
			\begin{tabular}{l |c}
                   \toprule
                    $\tau$ & Accuracy (\%) \\
                    \midrule
                    0.1  & 80.8\\
                    0.2  & \textbf{80.8 }\\
                     0.3  & 80.7\\
                     0.4 &80.3\\
                     0.5 & 80.1\\
                    \bottomrule
                \end{tabular}}
                  \label{stmix}
\end{table}

 \begin{table}
  \centering
  \small
  \caption{Complexity comparison of different methods. Iter-\textit{n} means performing the online saliency estimation per \textit{n} iterations.}
  \begin{tabular}{l|c|c|c}
   \toprule
   Method & Params & Training Time & Accuracy \\
    \midrule
    Baseline &1.46M & 2.85h & 74.4\\
    GLMC & 1.51M & 5.93h & 78.8\\
    ST-Mix (Ours) & 1.46M & 5.67h & 79.9 \\
    Shap-Mix Iter-1 (Ours) & 1.46M & 9.43h  & 80.8 \\ 
    Shap-Mix Iter-2 (Ours) & 1.46M & 8.11h & 80.6\\
    Shap-Mix Iter-5 (Ours) & 1.46M & 6.03h  & 80.7\\
    \bottomrule
\end{tabular}
  \label{tab:complex}
\end{table}

\subsection{Long-Tailed Action Recognition with Different Backbones}
Our method is compatible with different backbones. Here we present the results using different model backbones in Table~\ref{tab:backbones}. As we can see, our method can bring consistent performance improvement using different backbones, demonstrating the general effectiveness. Meanwhile, it is worth noting that different backbones can have significant performance differences under long-tailed recognition. Specifically, ST-GCN++, Info-GCN, and HD-GCN are less robust than the CTR-GCN in the long-tailed action recognition, although they can yield a better performance in the balanced setting. This means that the model architecture design is also important as the future work to handle the long-tailed skeleton-based action recognition.
 
\subsection{Ablation Study on the Temperature Coefficient}
Recall that we utilize the saliency maps as the importance distribution and sample the mixed parts from it. Specifically, the distribution is derived from the list of Shapley values:
\begin{equation}
    \centering
    d(c) = softmax(norm(\{v^c_b/|b|\})/\tau),
\end{equation}
where $\tau$ is the temperature hyper-parameter, controlling the sharpness of the obtained distribution. The effect of the temperature is demonstrated in Table~\ref{tab:temper}. As we can see, when the temperature is too high, the guidance is weak due to the smooth distribution. We set the temperature to 0.2, which gives the best performance.  

%\begin{figure}[t]
%    \centering
%    \includegraphics[width=0.40\textwidth]{figure/temper.pdf}
%   \caption{
%   Ablation study on the temperature on the LT-NTU 60 dataset.}
%  \label{fig:tem}
%\end{figure}

\subsection{Ablation Study on the Configurations of ST-Mix}
We conduct the ablation study under balanced settings on the hyper-parameters in spatial and temporal designs, including the body part number and the frame length to be mixed. As shown in Table~\ref{tab:spatial} and Table~\ref{tab:temporal}, the number of mixed parts and mixed frame length are important to the performance of the model. We choose the implementation giving the best performance. 
 
\subsection{Complexity Analysis of the Method}
The analysis of the complexity of the proposed method is provided in Table~\ref{tab:complex}. The training time is reported using a NVIDIA 4090 GPU for all models. As we can see, our method is based on augmentation and almost introduces no additional parameter overhead compared with the baseline algorithm. However, it is noted that the computational complexity has increased notably. The overhead comes from two aspects. 1) The first is the optimization of the mixed samples, which increases the batch size as shown in ST-Mix. 2) The second one is that the saliency estimation is baked into our training process, \ie, we perform the online saliency estimation in every iteration. Fortunately, we can easily extend our approach to perform online saliency estimation once every \textit{n} iterations to reduce the overhead brought by the online saliency estimation, making a good trade-off between estimation accuracy and computational complexity. Note that the inference time of these methods is the same because of the same backbone adopted.

\subsection{Visualization Results of the Shapley Value Guided Saliency Estimation}
\begin{figure*}[ht]
    \centering
    \includegraphics[width=0.98\textwidth]{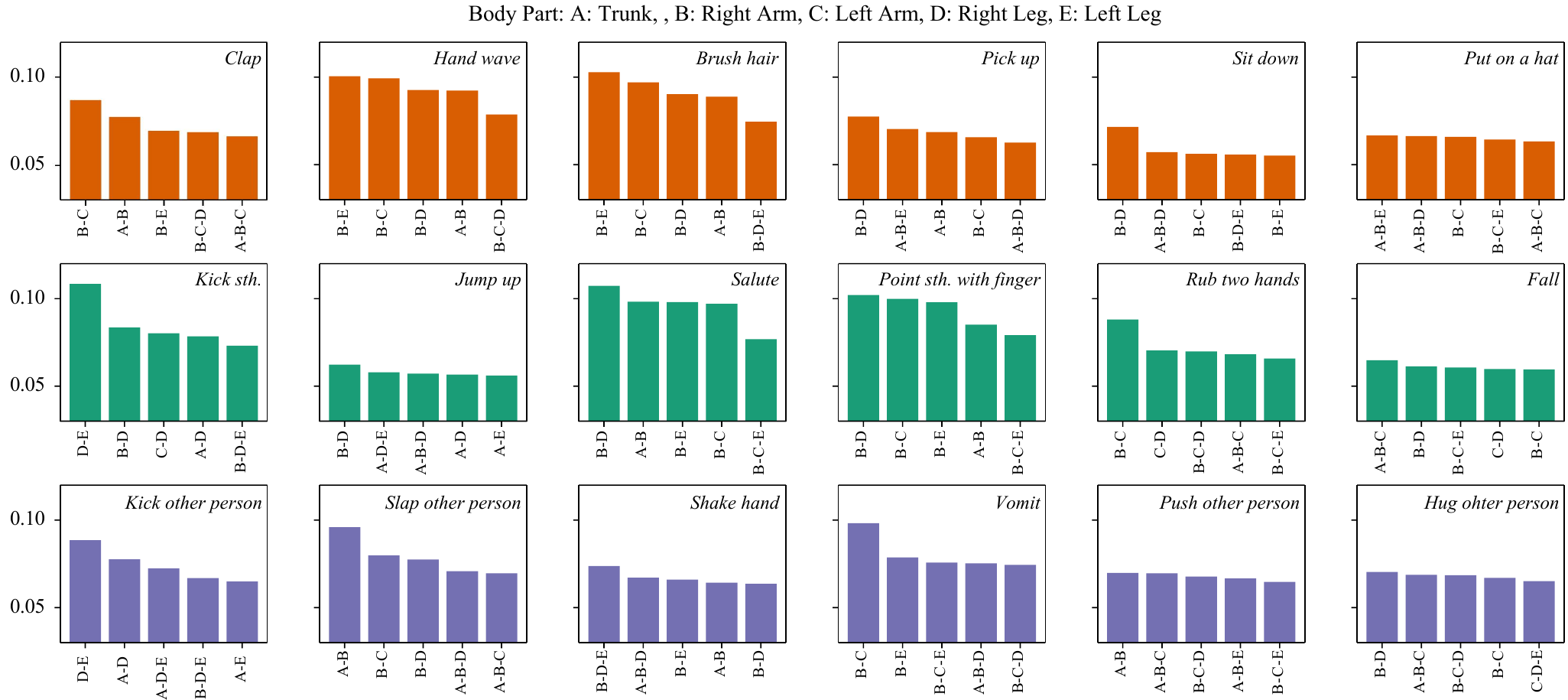}
   \caption{
   Visualization results of the Shapley value guided saliency estimation on LT-NTU 60 dataset. The first, second, and the third rows are the actions from many-, medium-, few-shot classes, respectively, where the top 5 most salient parts are given. Note that the Shapley value is normalized and the average saliency is 0.05. }
  \label{fig:shap}
\end{figure*}

We present more visualization results of saliency estimation in Figure~\ref{fig:shap}. As we can see, Shapley value can give a rational estimation result for different action categories owing to its superiority in considering the interaction of different skeleton parts. Although the estimation results can be more fine-grained if we divide the skeletons into more parts, it would result in a significant computational head. Meanwhile, considering that the main motion tend to occur in the middle of a sequence, we only perform the saliency estimation in the spatial dimension to make a trade-off between the estimation complexity and the granularity of saliency maps. 

We can find that most people's dominant hand, the right hand, is very prominent in most actions, which is intuitive. Meanwhile, in some actions involving the whole body parts, \eg, falling and jumping, the salience of different parts is relatively balanced, \ie, close to 0.05. This is because each body part participates in these actions and implies these actions as well.

\end{document}